\documentclass[letterpaper, 10 pt, conference]{ieeeconf}

\IEEEoverridecommandlockouts                              
\def\figurename{Fig. }

\usepackage[cmex10]{amsmath}
\usepackage{eufrak}
\usepackage{multirow}
\usepackage[caption=false,font=footnotesize]{subfig}
\usepackage[pdftex]{graphicx}
\DeclareGraphicsExtensions{.pdf,.jpeg,.png,.jpg}
\usepackage{cite}

\title{\LARGE \bf
Online Global Loop Closure Detection for Large-Scale Multi-Session Graph-Based SLAM
}

\author{Mathieu Labb\'e$^{1}$ and Fran{\c c}ois Michaud$^{1}$% <-this % stops a space
\thanks{*This work was supported by the Natural Sciences and Engineering Research Council of Canada (NSERC), the Canada Research Chair program and the Canadian Foundation for Innovation }% <-this % stops a space
\thanks{$^{1}$M. Labb\'e and F. Michaud are with the Interdisciplinary Institute of Technological Innovation (3IT) and the Department of Electrical Engineering and Computer Engineering, Universit\'{e} de Sherbrooke, 2500 boul. Universit\'{e}, Sherbrooke, Qu\'{e}bec, CANADA {\tt\small \{Mathieu.M.Labbe, Francois.Michaud\}@USherbrooke.ca} }
}

\begin{document}

\maketitle
\thispagestyle{empty}
\pagestyle{empty}

%%%%%%%%%%%%%%%%%%%%%%%%%%%%%%%%%%%%%%%%%%%%%%%%%%%%%%%%%%%%%%%%%%%%%%%%%%%%%%%%
\begin{abstract}
For large-scale and long-term simultaneous localization and mapping (SLAM), a robot has to deal with unknown initial positioning caused by either the kidnapped robot problem or multi-session mapping. This paper addresses these problems by tying the SLAM system with a global loop closure detection approach, which intrinsically handles these situations. However, online processing for global loop closure detection approaches is generally influenced by the size of the environment. The proposed graph-based SLAM system uses a memory management approach that only consider portions of the map to satisfy online processing requirements. The approach is tested and demonstrated using five indoor mapping sessions of a building using a robot equipped with a laser rangefinder and a Kinect.
\end{abstract}

%%%%%%%%%%%%%%%%%%%%%%%%%%%%%%%%%%%%%%%%%%%%%%%%%%%%%%%%%%%%%%%%%%%%%%%%%%%%%%%%
\section{INTRODUCTION}
Autonomous robots operating in real life settings must be able to navigate in large, unstructured, dynamic and unknown spaces. To do so, they must build a map of their operating environment in order to localize itself in it, a problem known as Simultaneous localization and mapping (SLAM). A key feature in SLAM is detecting previously visited areas to reduce map errors, a process known as loop closure detection. Our interest lies with graph-based SLAM approaches \cite{lu1997globally} that use nodes as poses and links as odometry and loop closure transformations.

While single session graph-based SLAM has been largely addressed \cite{Bosse04, thrun2006graph, grisetti2010tutorial}, multi-session SLAM involves having to deal with the fact that robots, over a long period of operation, will eventually be shutdown and moved to another location without knowing it. Such situations include the so-called kidnapped robot problem and the initial state problem: when it is turned on, a robot does not know its relative position to a map previously created. One way to do multi-session mapping is to have the robot, on startup, localize itself in a previously-built map. This solution has the advantage to always use the same referential and only one map is created across the sessions. However, the robot must start in a portion of the environment already mapped, otherwise it never can relocalize itself in it. Another approach is to initialize a new map with its own referential and when a previously visited location is encountered, the transformation between the two maps can be computed. In \cite{mcdonald2012real}, special nodes called ``anchor nodes" are used to keep transformation information between the maps. A similar approach is also used with multi-robot mapping \cite{kim2010multiple}: transformations between maps are computed when a robot sees the other or when a landmark is seen by both robots in their respective maps.

Global loop closure detection approaches, by being independent of the robot's estimated position \cite{Ho06}, can intrinsically solve the problem of determining when a robot comes back to a previous map using a different referential \cite{cummins2011appearance}. Popular global loop detection approaches are appearance-based \cite{Angeli08c, botterill2011bag, konolige2010view, booij2009efficient}, exploiting the distinctiveness of images. The underlying idea behind these approaches is that loop closure detection is done by comparing all previous images with the new one. When loop closures are found between the maps, a global graph can be created by combining the graphs from each session. Graph pose optimization approaches \cite{Folkesson07, Grisetti07a, johannsson2012temporally} can then be used to reduce odometry errors using poses and link transformations inside each map and also between the maps. 

All the solutions above can be integrated together to create a functional graph-based SLAM system. However, for loop closure detection and graph optimization approaches, online constraint satisfaction is limited by the size of the environment. For large-scale and long-term operation, the bigger the map is, the more computing power is required to process the data online. Mobile robots have limited computing resources, therefore online map updating is limited, and so some parts of the map must be somewhat forgotten. Memory management approaches \cite{labbe13appearance} can be used to limit the size of the map so that loop closure detections are always processed under a fixed time limit, thus satisfying online requirements for long-term and large-scale environment mapping. 

The solution presented in this paper simultaneously addresses these two problems: multi-session mapping, and online map updating with limited computing resources. Global loop closure detection is used across the mapping sessions to detect when the robot revisits a previous map. Using these loop closure constraints, the graph is optimized to minimize trajectory errors and to merge the maps together in the same referential. A memory management mechanism is used to limit the data processed by global loop closure detection and graph optimization in order to respect online constraints independently of the size of the environment. The algorithm is tested over five mapping sessions using a robot in an indoor environment. 

The paper is organized as follows. Section \ref{sec:system_description} describes our approach. Section \ref{sec:results} presents experimental results and Section \ref{sec:discussion} discusses limitations of the approach on very long-term operation. Section \ref{sec:conclusion} concludes the paper.

%%%%%%%%%%%%%%%%%%%%%%%%%%%%%%%%%%%%%%%%%%%%%%%%%%%%%%%%%%%%%%%%%%%%%%%%%%%%%%%%

%%%%%%%%%%%%%%%%%%%%%%%%%%%%%%%%%%%%%%%%%%%%%%%%%%%%%%%%%%%%%%%%%%%%%%%%%%%%%%%%
\section{ONLINE MULTI-SESSION GRAPH-BASED SLAM}
\label{sec:system_description}

In our approach, the underlying structure of the map is a graph with nodes and links. The nodes save odometry poses for each location in the map. The nodes also contain visualization information like laser scans, RGB images, depth images and visual words \cite{sivic2003video} used for loop closure detection. The links store rigid geometrical transformations between nodes. There are two types of links: neighbor and loop closure. Neighbor links are added between the current and the previous nodes with their odometry transformation. Loop closure links are added when a loop closure detection is found between the current node and one from the same or previous maps. 
Our contribution in this paper involves combining two algorithms, loop closure detection \cite{labbe13appearance} and graph optimization \cite{Grisetti07a}, through a memory management process \cite{labbe13appearance} that limits the number of nodes available from the graph for loop closure detection and graph optimization, so that they always satisfy online requirements. 

\subsection{Loop Closure Detection}
\label{loopclosuredetection}
For global loop closure detection, the bag-of-words approach described in \cite{labbe13appearance} is used. Briefly, this approach uses a bayesian filter to evaluate loop closure hypotheses over all previous images. When a loop closure hypothesis reaches a pre-defined threshold $H$, a loop closure is detected. Visual words, which are SURF features quantized to an incremental visual dictionary, are used to compute the likelihood required by the filter. 

In this paper, the RGB image, from which the visual words are extracted, is registered with a depth image, i.e., for each 2D point in the RGB image, a 3D position can be computed using the calibration matrix and the depth information given by the depth image. The 3D positions of the visual words are then known. When a loop closure is detected, the rigid transformation between the matching images is computed by a RANSAC approach using the 3D visual word correspondences. If a minimum of $I$ inliers are found, loop closure is accepted and a link with this transformation between the current node and the loop closure hypothesis node is added to the graph. If the robot is constrained to operate on a single plane, the transformation can be refined with 2D iterative-closest-point (ICP) optimization \cite{besl1992method} using laser scans contained in the matching nodes.

\subsection{Graph Optimization}\label{sec:graphoptimization}
TORO \cite{Grisetti07a} (Tree-based netwORk Optimizer) is the graph optimization approach used, in which node poses and the link transformations are used as constraints. When loop closures are found, the errors introduced by the odometry can then be propagated to all links, thus correcting the map. It is relatively straightforward to use TORO to create a tree from the map's graph when there is only one map: the TORO tree has therefore only one root. In multi-session mapping, the different maps created have their own root with their own reference frames. When loop closures occur between the maps, TORO cannot optimize the graph if there are multiple roots. It may also be difficult to find a unique root if some portions of the map are forgotten or unavailable at that time (because of the memory management approach used to satisfy online processing requirements, explained in Sect. \ref{rtab}). 
%FM: Petits ajustements avant et après
To alienate these problems, our approach takes the root of the tree to be the latest node added to the current map graph, which is always uniquely defined across intra-session and inter-session mapping.

\subsection{Memory Management for Online Multi-Session Mapping}\label{rtab}
For online mapping, new incoming data must be processed faster than the time required to acquire them. For example, if data are acquired at 1 Hz, new data should be added to the graph with global loop closure detection and graph optimization should be done in less than $R=1$ second. The problem is that the time required for loop closure detection and graph optimization depends on the map's graph size. Long-term and large-scale online mapping is then limited by the size of the environment. 
To handle this, the RTAB-Map memory management approach \cite{labbe13appearance} is used to maintain a graph manageable online by the loop closure detection and graph optimization algorithms, thus making the metric SLAM approach presented in this paper independent of the size of the environment. 

\begin{figure}[!t] 
\centering 
\includegraphics[width= 2.3in]{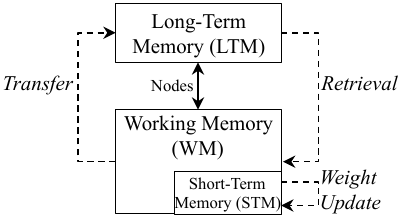} 
\caption{Memory management model.} 
\label{fig:memory} 
\end{figure}

The approach works as follows. The memory is composed of a Short-Term Memory (STM), a Working Memory (WM) and a Long-Term Memory (LTM), as shown by Figure \ref{fig:memory}. The STM is the entry point for new nodes added to the graph when new data are acquired, and has a fixed size $S$. Nodes in STM are not considered for loop closure detection because they are generally very similar from one to another. When the STM size reaches $S$ nodes, the oldest node is moved to WM to be considered for loop closure detection. The WM size indirectly depends on a fixed time limit $T$. When the time required to process the new data reaches $T$, some nodes of the graph are transferred from WM to LTM, thus keeping the WM size nearly constant. The LTM is not used for loop closure detection and graph optimization. However, if a loop closure is detected, neighbors in LTM of the old node can be transferred back to WM (a process called Retrieval) for further loop closure detections. In other words, when a robot revisits an area which was previously forgotten, it can remember incrementally the area if a least one node of this area is still in WM. 

The choice of which nodes to keep in WM is based on a Weight Update step done in STM. The heuristic used to increase the weight of a node is based on the principle that, as humans do  \cite{atkinson1968human, baddeley1997human}, the robot should remember more the areas where they spent most of their time in. Therefore, the longer the robot is at a particular location, the larger the weight of the node should be. 
If two consecutive images are similar, i.e., the ratio of corresponding visual words between the images is over a specified threshold $Y$, the node's weight of the first image is increased by one and no new node is created for the second image. 
By following this heuristic, the compromise made between search time and space is therefore driven by the environment and the experiences of the robot. Oldest and less weighted nodes in WM are transferred to LTM before the others, thus keeping in WM only the nodes seen for longer periods of time.

\begin{figure}[!t] 
\centering 
\includegraphics[width= 2.3in]{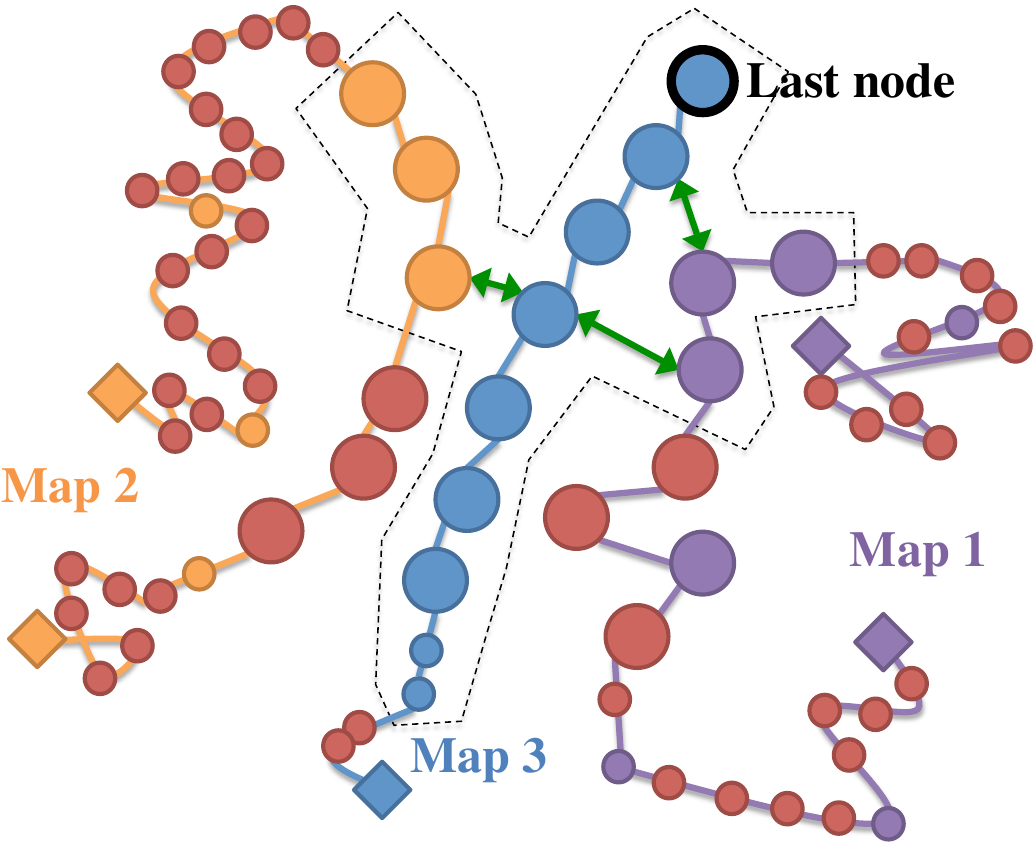} 
\caption{Illustration of a local map created from multi-session mapping.} 
\label{fig:tree} 
\end{figure}

For the approach presented in this paper, a local map consists of the biggest fully connected graph that can be created through neighbor and loop closure links from the last node (used as the root) with those in WM. Figure \ref{fig:tree} illustrates the concept. The diamonds represent initial and end nodes for each mapping session. The nodes in LTM are shown in red and the others are those in WM. The current local map is created and optimized only using nodes in WM that are linked to the last node (all nodes in the dashed area). The local map therefore represents more than the latest mapping session: it can span over multi-session mapping through loop closure links (green links). The other nodes still in WM that are not included in the local map are unreachable from the last node through links available in WM at this time.

Using this memory management approach, some parts of the map may be missing for graph optimization, as described in 
\ref{sec:graphoptimization}. Online graph optimization is done on the local map, with the constraints available in WM at that time. Constraints transferred to LTM are not used, thus limiting graph quality compared to using all constraints available. This is the compromise to make to be able to satisfy online processing requirements. However, if required, the approach is 
still able to create a global map by using all constraints from LTM and conduct offline a global graph optimization.

%%%%%%%%%%%%%%%%%%%%%%%%%%%%%%%%%%%%%%%%%%%%%%%%%%%%%%%%%%%%%%%%%%%%%%%%%%%%%%%%
\section{RESULTS}
\label{sec:results}
\label{results}

\begin{figure}[!t] 
\centering 
\includegraphics[width= 1.9in]{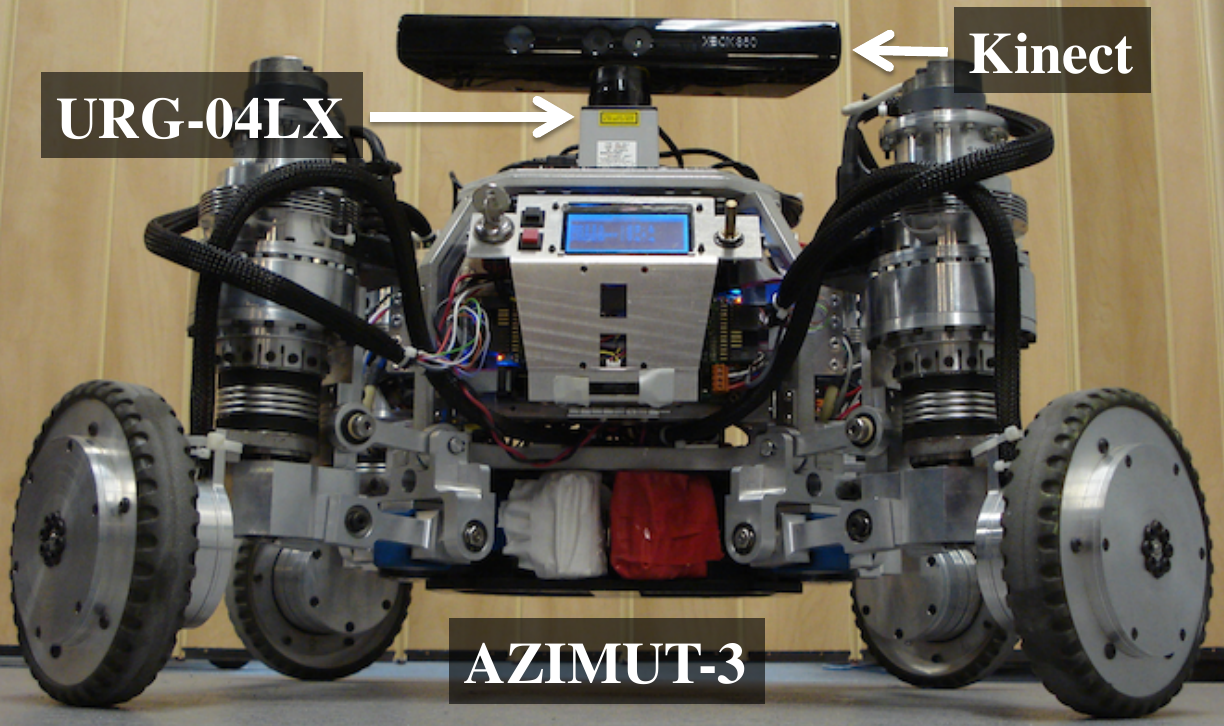} 
\caption{AZIMUT-3 robot equipped with a URG-04XL laser range finder and a Kinect sensor.} 
\label{fig:azimut} 
\end{figure}

The data sets used for the experiments are acquired using the AZIMUT-3 robot \cite{ferland10teleopration}, shown by \figurename \ref{fig:azimut}, equipped with a URG-04LX laser rangefinder and a Kinect sensor. The RGB images from the Kinect are used for the appearance-based loop closure detection while the depth images are used to find the 3D position of the visual words. Laser scans and RGB-D point clouds created from the Kinect are used for map visualization. As mentioned in \ref{loopclosuredetection}, since in this experiment the robot is constrained to a single plane, loop closure transformations 
are refined using 2D ICP with the laser scans to increase precision: the transformations are then limited to three degrees of freedom ($x$, $y$ and rotation over $z$ axis), ignoring noise on other degrees of freedom computed by the visual transformation.

Five mapping sessions (total length of 750 m) were conducted by starting the robot at different locations in our lab building. Between the mapping sessions, the robot was turned off to reset odometry, and moved to another location. In each session, the robot revisited at least one part of the environment mapped in a previous session. Data acquisition is done using the ROS bag mechanism (http://ros.org). 
Odometry, laser scans, RGB images and depth images are recorded at 1 Hz 
(i.e., $R=1$ s) in a ROS bag. A ROS bag can be played using the same timings as during acquisition, making a realistic input for mapping and a good common format for other algorithms using ROS. One ROS bag per mapping session is taken. The ROS bags are processed %FM
on a MacBook Pro 2010: 2.66 GHz Intel Core i7 and SSD hard drive (on which the LTM is saved).

\begin{figure}[!t] 
\centering 
\includegraphics[width= 2.3in]{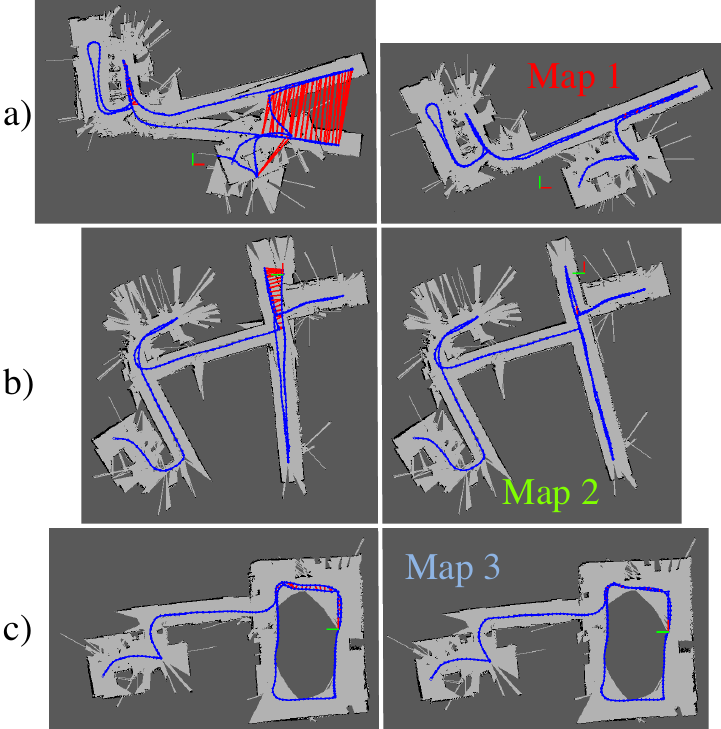} 
\caption{Resulting local maps without (left) and with (right) graph optimizations for a) Map 1, b) Map 2 and c) Map 3. Loop closures are shown in red.} 
\label{fig:independentMap123} 
\end{figure}

\begin{figure}[!t] 
\centering 
\includegraphics[width= 3in]{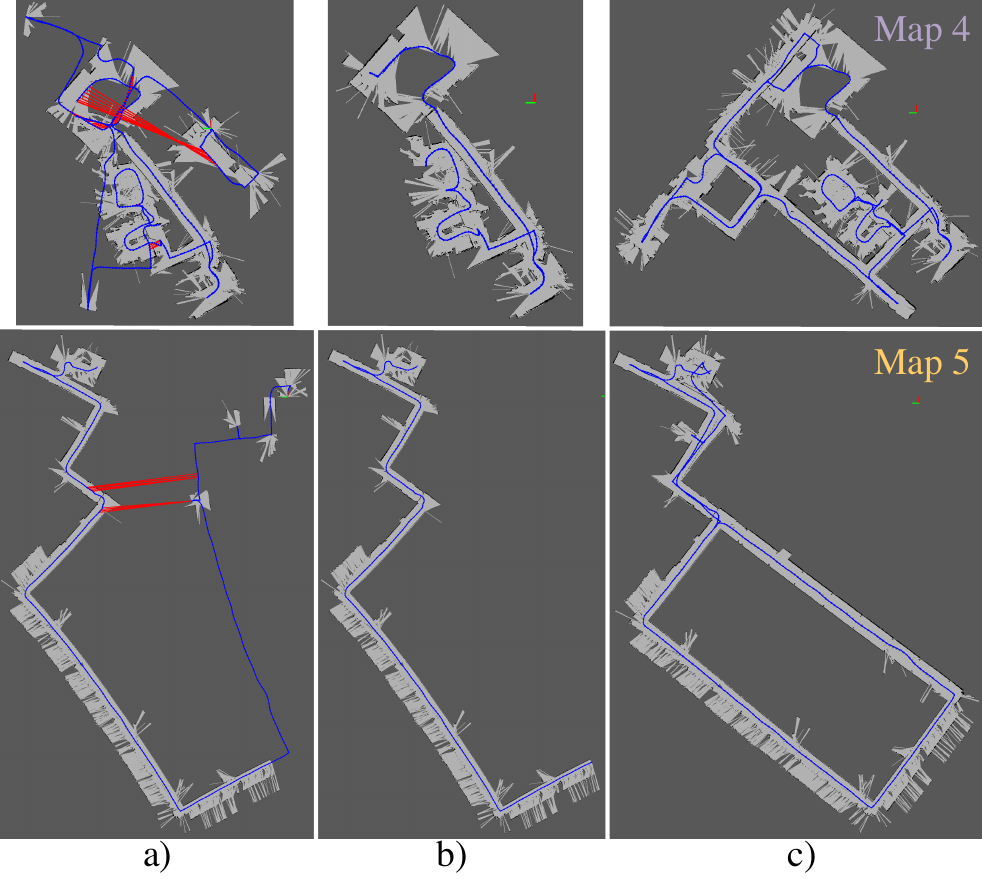} 
\caption{Results for Map 4 (top) and Map 5 (bottom), with a) the map from all nodes still in WM (light gray) with the global graph (blue line) not optimized, b) the local map with local graph optimization and c) the global map with global graph optimization. Loop closures are shown in red.} 
\label{fig:independentMap45} 
\end{figure}

\begin{figure}[!t] 
\centering 
\includegraphics[width= 2.6in]{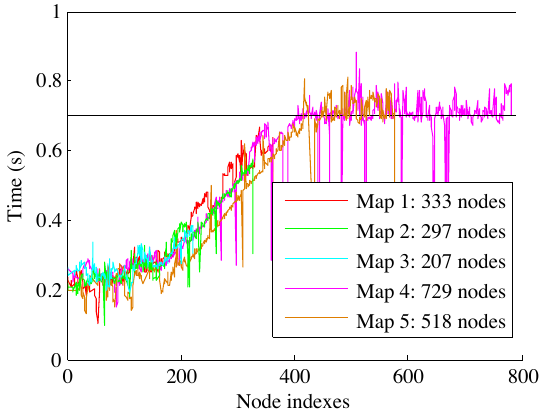} 
\caption{Processing time in relation to the number of nodes processed over time for each data set. $T$ is shown by the horizontal line.} 
\label{fig:timeIndependent} 
\end{figure}

Two experiments were conducted (STM size $S=10$, minimum inliers $I=5$ of RANSAC, hypothesis threshold $H=0.11$ and similarity threshold $Y=0.45$). For the first experiment, our approach processed each mapping session independently, i.e., the memory was cleared between each session. Time limit $T$ was set to $0.7$ s. \figurename \ref{fig:independentMap123} shows the resulting maps for sessions 1, 2 and 3, with and without graph optimizations. The light gray areas are empty spaces detected using the laser rangefinder. No nodes were transferred to LTM in these experiments (local maps are equal to global maps). This is confirmed by \figurename \ref{fig:timeIndependent}: $T$ was never reached for these sessions, and thus all nodes were used for loop closure detection and graph optimization. 
\figurename \ref{fig:independentMap45} shows results for the mapping sessions 4 and 5
(i.e., Map 4 and Map 5): the global graph not optimized (left), the last local map (middle) and the global map (right). The local map is the biggest map that was created online from the last node (with nodes available in WM), and the global map was generated offline after the mapping sessions (with all nodes in WM and LTM). As shown by \figurename \ref{fig:timeIndependent}, $T$ was reached before the end. \figurename \ref{fig:independentMap45} b) illustrate the effect of transferring nodes to LTM to satisfy the online requirement. Even if loop closures can be detected with older portions of the map still in WM (as shown in a)), the maps cannot be globally optimized if the neighbors of the loop closures are in LTM. For comparison, \figurename \ref{fig:independentMap45} c) are maps created offline using all constraints in LTM: here, loop closures with old portions of the map have an effect on graph optimization. 

\begin{figure}[!t] 
\centering 
\includegraphics[width= 2.4in]{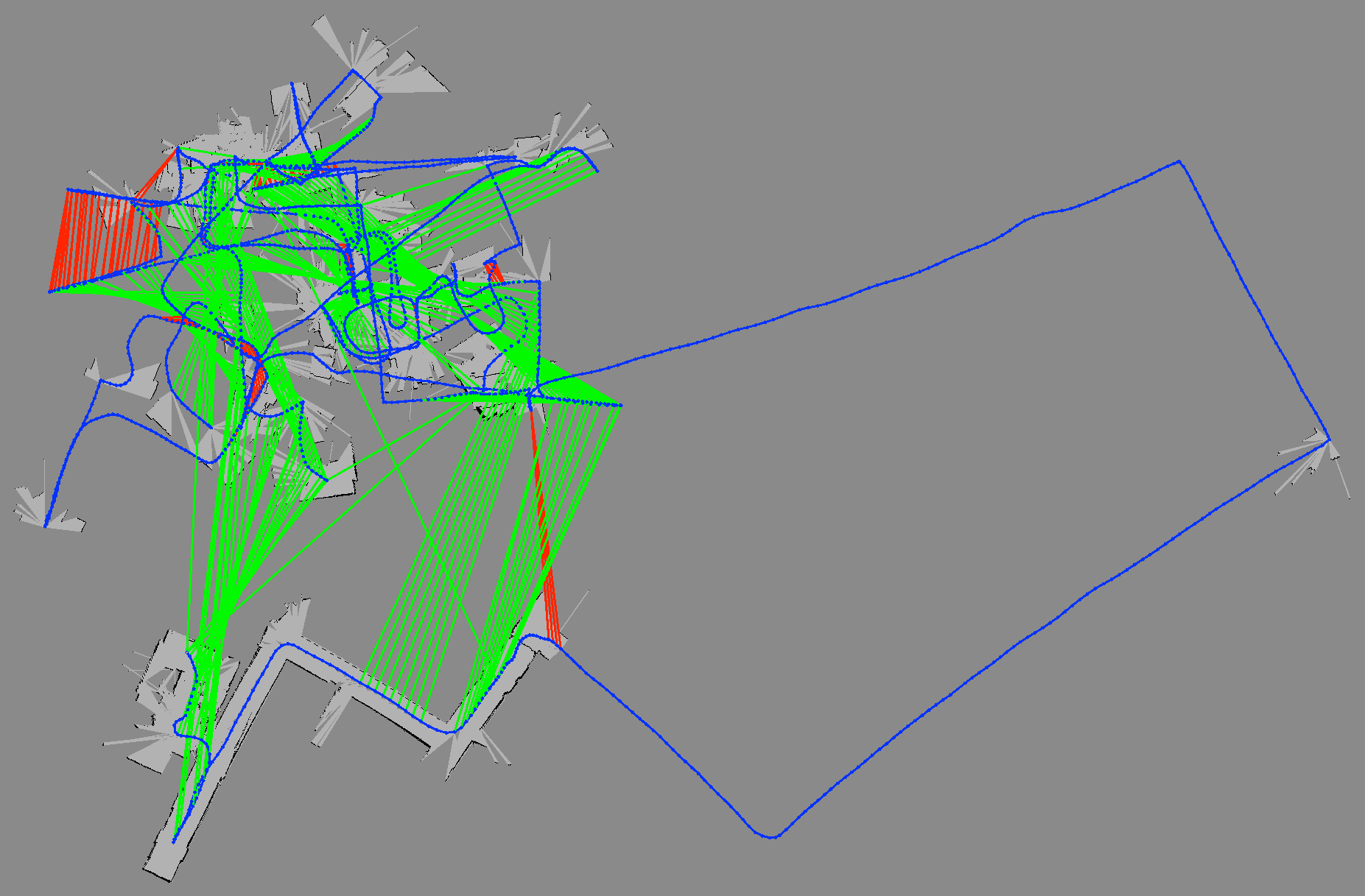} 
\caption{Top view of the map without optimization after five mapping sessions. The red and green links show intra-session and inter-session loop closures detected, respectively.} 
\label{fig:raw700} 
\end{figure}

\begin{figure}[!t] 
\centering 
\includegraphics[width= 2.7in]{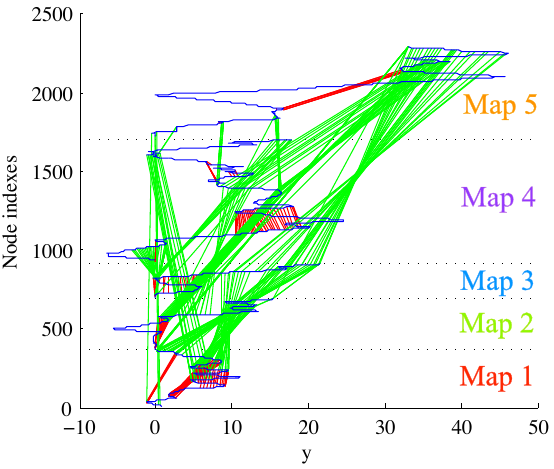} 
\caption{Loop closures between the mapping sessions. Only the $y$ values of the poses are illustrated for visibility purposes. Green and red links are inter-session and intra-session loop closures detected, respectively. Neighbor links are shown in blue. Note that only green links connect the five maps together.} 
\label{fig:loops} 
\end{figure}

\begin{figure}[!t] 
\centering 
\includegraphics[width= 2.7in]{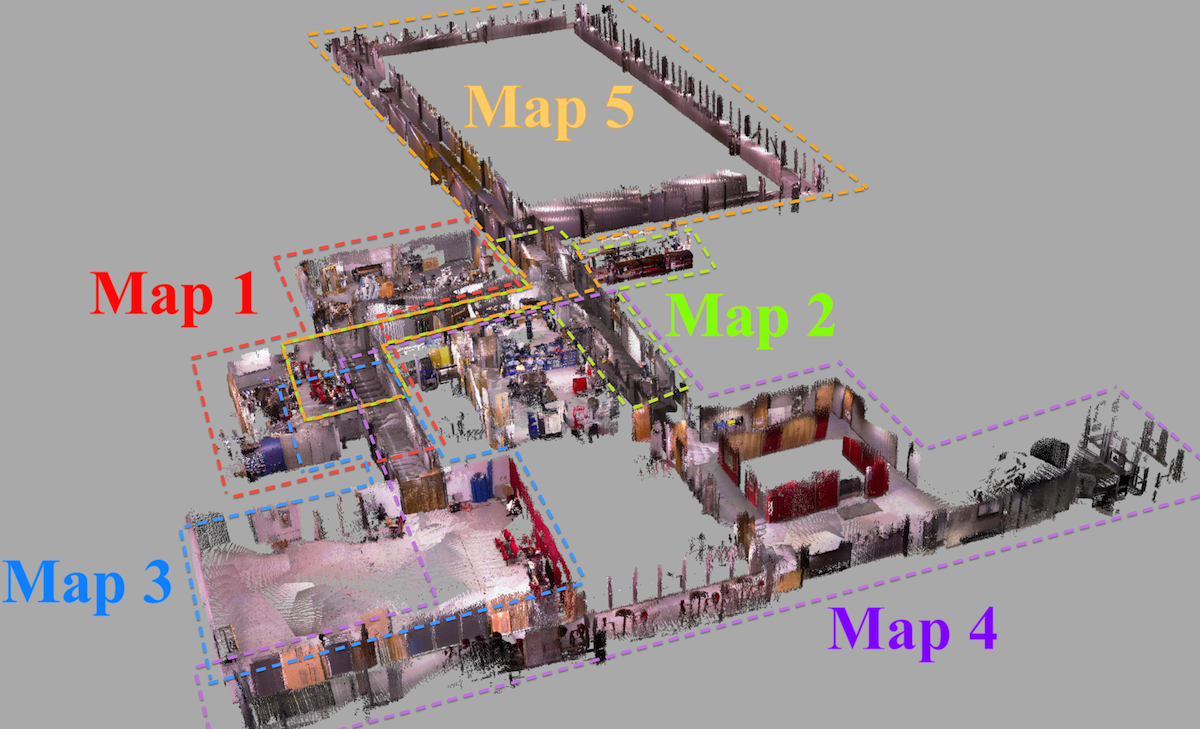} 
\caption{Five online mapping sessions merged together automatically.} 
\label{fig:3dareatexthd} 
\end{figure}

For the second experiment, the data sets for the five maps were processed one after each other, as in a real multi-session mapping trial. The robot automatically started a new map when the odometry was reset to zero before each session. The memory was preserved between the sessions and $T$ was also set to $0.7$ s.  \figurename \ref{fig:raw700} shows the last local map (nodes in light gray areas are those in WM) and global graph (blue line) without optimization.  
The maps lie over each other because they are all starting from the same referential. Loop closures detected in the same map (intra-session) and those detected between the maps (inter-session) are shown in red and green, respectively. To distinguish more easily inter-session loop closures, \figurename \ref{fig:loops} illustrates the global graph for $y$-value of the poses over time. Note that all paths for each session started at $y=0$ and they were not connected together by neighbor links. Optimizing the graph using all these detected loop closures results in a single fully connected map of all five mapping sessions. \figurename \ref{fig:3dareatexthd} shows the resulting global map by assembling the RGB-D point clouds from the Kinect using the optimized poses of the graph. 

\begin{figure}[!t] 
\centering 
\includegraphics[width= 2.9in]{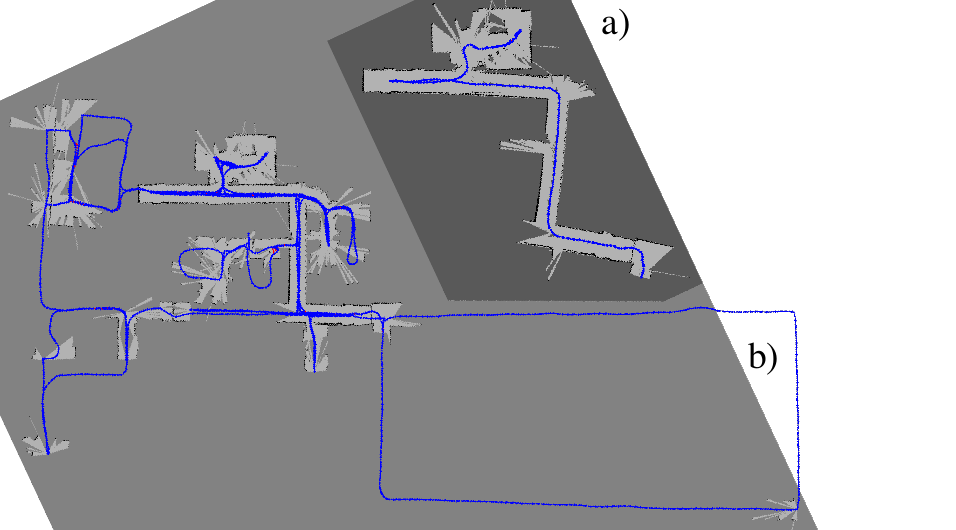} 
\caption{Graphs optimized for a) the last local map built online, b) the global map built offline, with nodes in light gray areas are those still in WM, and the other nodes are in LTM.} 
\label{fig:currentMapAndWM} 
\end{figure}

\figurename \ref{fig:currentMapAndWM} a) shows the resulting local map created from all the mapping sessions. Because the local map is built only from nodes in WM that are linked (directly or indirectly) to the last node, only a small portion of the global map is available online. Note that the local map is also smaller than Map 5 taken independently (shown by \figurename \ref{fig:independentMap45}): in the second experiment, there were nodes with more weight from previous mapping sessions that were still in WM, thus more nodes from the latest mapping session were transferred to LTM and not used for local map creation. These high weighted nodes are located in the light gray areas of \figurename \ref{fig:currentMapAndWM} b). The blue line represents the global graph created using all constraints in LTM. When using all constraints in LTM, the local map is also slightly more straight. At the end of the experiment, the global graph has 2074 nodes with all mapping sessions connected, with 330 nodes in WM (107, 12, 27, 28, 156 nodes from maps 1, 2, 3, 4 and 5, respectively) for which 173 nodes are accessible for the local map (4, 15, 6, 0, 148 nodes from maps 1, 2, 3, 4 and 5, respectively). For the local map, it is normal that a high proportion of nodes are from the last session, which is the most recent one. Nodes from older maps are those retrieved from LTM around the latest loop closures found. For example, when the robot is mapping a new area, only nodes of the last session would be in the local map.

\begin{figure}[!t] 
\centering 
\includegraphics[width= \columnwidth]{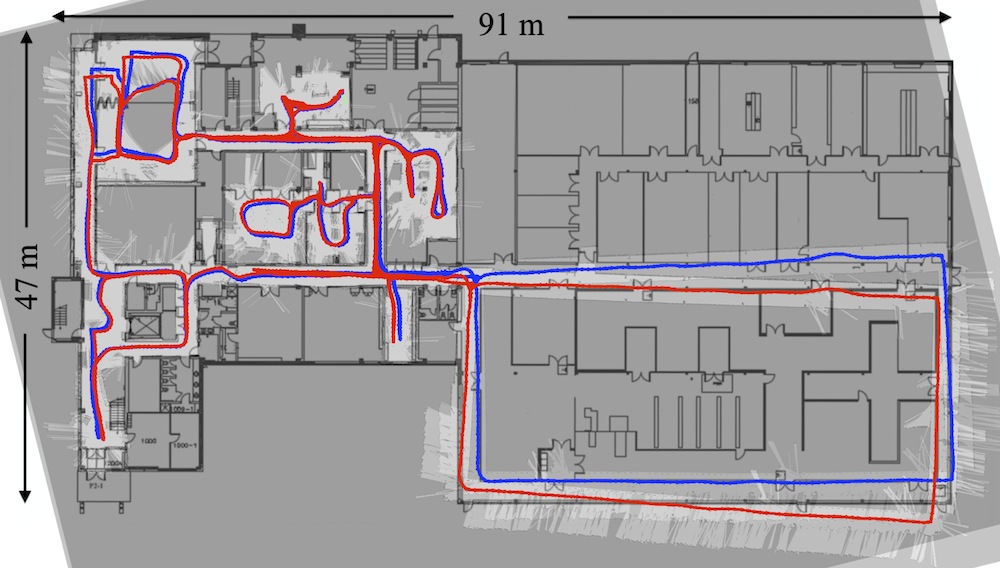} 
\caption{Global maps with (blue) and without (red) $T$. The maps are manually superimposed over the actual plan of the building.} 
\label{fig:mixed} 
\end{figure}

\begin{figure}[!t] 
\centering 
\includegraphics[width= 2.4in]{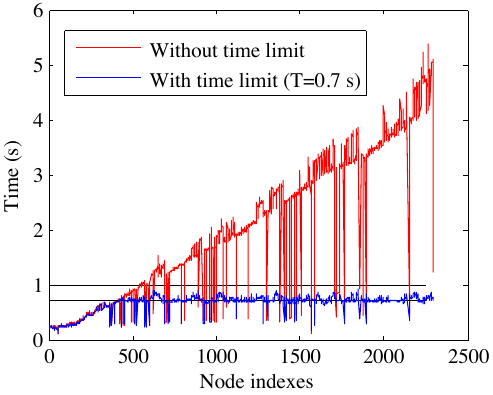} 
\caption{Processing time for each node added to graph. The horizontal lines are $T=0.7$ and $R=1$.} 
\label{fig:timeAll} 
\end{figure}

To observe the influence of memory management on the quality of the map created, we conducted the same experiment without $T$. All nodes were then kept in WM and they were processed by both loop closure detection and graph optimization at each time step. Normally, without transferring nodes to LTM, more loop closures would be detected, so more constraints would be used for graph optimization. As shown in \figurename \ref{fig:timeAll}, the processing time becomes greater than the acquisition time $R$, which is not the case with $T=0.7$ s. However, without $T$, 193 intra-session and 387 inter-session loop closures were detected, comparatively to 188 and 258 respectively for the online experiment. \figurename \ref{fig:mixed} compares the resulting global maps with (blue) and without (red) $T$. 
By comparing with the building plan (the plan was scaled to 5 cm / pixel like the generated maps, the maps were manually oriented so trajectories are aligned to most doors traversed), the quality of the experiment without $T$ (red) is a little better than with $T$ (blue), probably because more loop closures were used for graph optimization. However, for the two conditions, the large loop from Map 5 is not correctly aligned with the building plan. The robot traversed this area only once and exited from the same door from which it entered, making it more difficult for the graph optimization algorithm to correct angular errors for this single entry point. For comparison, the left part of the map was also traversed once during session 4, but the robot exited the area from another door, thus making the area more robust to angular errors. 

%%%%%%%%%%%%%%%%%%%%%%%%%%%%%%%%%%%%%%%%%%%%%%%%%%%%%%%%%%%%%%%%%%%%%%%%%%%%%%%%
\section{DISCUSSION}
\label{sec:discussion}

In term of processing time, the results show that the proposed approach is able to satisfy online processing requirements independently of the size of the environment. However, map quality depends on the number of loop closures that can be detected. To satisfy online requirements, the robot transfers in LTM some portions of the map which cannot be used for loop closure detection. For multi-session mapping, the worst case would occur if all nodes of a previous map are transferred to LTM before a loop closure is detected with the new map. This would result in definitely forgetting the previous map: there would be no links in WM and even in LTM that could connect this older map to the new one, and it would be ignored even for the global map construction. To avoid this problem, our approach could keep at least one node for each map in WM. However, if the number of mapping sessions becomes very high (e.g., thousands of sessions), these nodes would definitely have to be transferred in LTM to satisfy the online requirement. For long-term, large-scale and multi-session mapping, some portions of the map would then be definitely forgotten, and therefore some kind of heuristic to efficiently manage important nodes to keep in WM is required. 

Another observation is that frequently revisiting old maps increases global map quality. A robot autonomously mapping a facility could, when detecting an old map, decide to revisit some parts of it to detect more inter-session loop closures, thus creating more constraints for graph optimization.

In the experiments conducted, no invalid loop closures were detected. If this occur, erroneous constraints would be added to graph optimization, resulting in map errors. Some graph optimization approaches such as \cite{latif2012robust, sunderhauf2012towards} deal with possible invalid matches, and could be used to increase robustness of the proposed approach.

%%%%%%%%%%%%%%%%%%%%%%%%%%%%%%%%%%%%%%%%%%%%%%%%%%%%%%%%%%%%%%%%%%%%%%%%%%%%%%%%
\section{CONCLUSION}
\label{sec:conclusion}
Results presented in this paper suggest that the proposed graph-based SLAM approach is able to meet online requirements needed for large-scale, long-term and multi-session online mapping. By limiting the number of nodes in WM available for global loop closure detection and graph optimization, online processing is achieved for new data acquired. 
Our approach is tightly based on global loop closure detection, allowing it to naturally deal with the kidnapped robot problem and gross errors in odometry. 
Our code is open source and available at http://rtabmap.googlecode.com/. In future work, we plan to study the impact of autonomous exploration strategies on multi-session mapping, especially how it can actively direct exploration based on nodes available for online mapping and graph optimization.

%%%%%%%%%%%%%%%%%%%%%%%%%%%%%%%%%%%%%%%%%%%%%%%%%%%%%%%%%%%%%%%%%%%%%%%%%%%%%%%%

\bibliographystyle{IEEEtran}

\bibliography{IEEEabrv,../../../../Papers/References}

\end{document}